\documentclass[letterpaper,10pt,conference]{ieeeconf}  %

\usepackage[most]{tcolorbox}

\usepackage[utf8]{inputenc} %
\usepackage[T1]{fontenc}    %
\usepackage{url}            %
\usepackage{booktabs}       % 
\usepackage{nicefrac}       %
\usepackage{microtype}      %

\usepackage{colortbl}
\usepackage{graphicx}
\usepackage{amsmath,amsfonts,amssymb}
\usepackage{color}

\usepackage{xspace}
\usepackage{array}
\usepackage{cite}
\usepackage{xcolor}

\usepackage{wrapfig}
\usepackage{multirow}
\usepackage{adjustbox}
\usepackage{pifont}

\def\diff{\mathrm{d}}

\RequirePackage{mathtools}
\colorlet{green}{green!30!}
\colorlet{blue}{blue!30!}
\colorlet{red}{red!30!}
\colorlet{violet}{violet!30!}
\tcbset{
    inline/.style={
        on line, % Aligns the box with the baseline of the text
        arc=2pt, % Rounded corners with a small radius
        boxsep=2pt, % Padding around the content inside the box
        left=0pt, right=0pt, % Padding on the left and right sides
        top=0pt, bottom=0pt, % Minimal padding on the top and bottom
        boxrule=0pt, % No border
        colback=#1, % Lighter shade of the specified color
        colframe=#1, % Frame color same as the background (can be adjusted)
        coltext=black, % Text color
    }
}
\newcommand{\mathbox}[2][]{%
  \tcbox[inline=#1]{$#2$}
}

\newcommand{\boldparagraph}[1]{\vspace{0.0cm}\noindent{\bf #1.} }
\newcommand{\fadedtext}[1]{\textcolor{gray}{#1}}

\usepackage[font=small, labelfont=bf]{caption}
\usepackage{todonotes}

\usepackage{algorithm} %for algorithm environment
\usepackage{algpseudocode} %for algorithm environment

\newcommand{\cmark}{\ding{51}}%
\newcommand{\xmark}{\ding{55}}%
\usepackage{hyperref} 
\IEEEoverridecommandlockouts                              % This command is only needed if 
\usepackage{graphics} % for pdf, bitmapped graphics files

\title{\LARGE \bf
DualAD: Dual-Layer Planning for Reasoning in Autonomous Driving}

\author{ Dingrui Wang$^{1*}$, Marc Kaufeld$^{1*}$, Johannes Betz$^{1}$ %\\
\thanks{$^{1}$D. Wang, M. Kaufeld and J. Betz are with the Professorship of Autonomous Vehicle Systems, TUM School of Engineering and Design, Technical University Munich, 85748 Garching, Germany; Munich Institute of Robotics and Machine Intelligence (MIRMI), \{{dingrui.wang, marc.kaufeld, johannes.betz}\}@tum.de
}
\thanks{$^{*}$Equal contribution.}
}% <-this % stops a space

\begin{document}

\makeatletter
\let\@oldmaketitle\@maketitle%
\renewcommand{\@maketitle}{\@oldmaketitle%
    \centering

    \includegraphics[width=\textwidth]{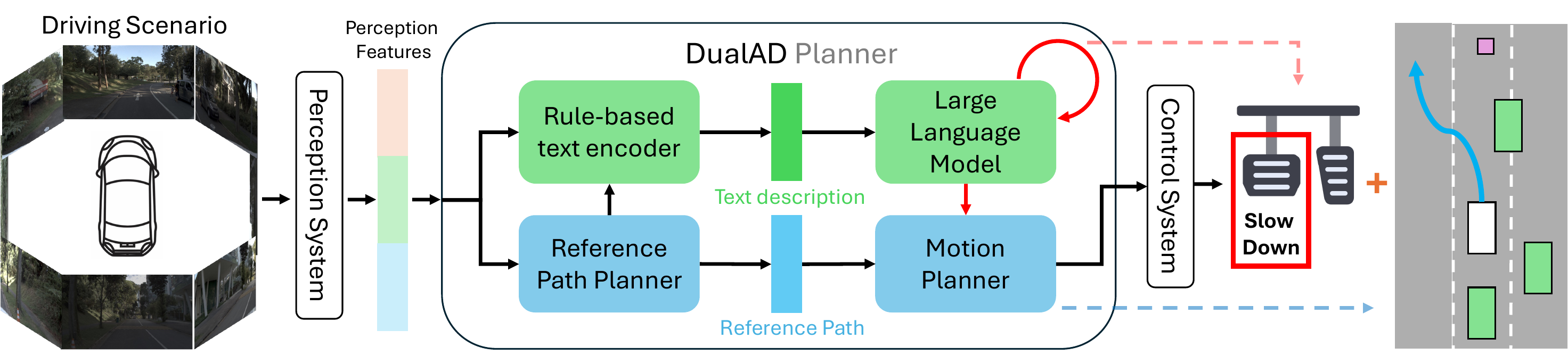}
    \captionof{figure}{DualAD is a dual-layer autonomous driving framework that imitates human cognitive processes during driving. The lower layer is responsible for reference path and motion planning, while the upper layer is the the reasoning module which dynamically checks the surrounding potential danger and adjusts speed limits or even applies hard braking in critical scenarios.}
    \label{fig:teaser}
    \vspace*{-3.8mm}
}
\makeatother
\maketitle

\setcounter{figure}{1}

\maketitle

%%%%%%%%%%%%%%%%%%%%%%%%%%%%%%%%%%%%%%%%%%%%%%%%%%%%%%%%%%%%%%%%%%%%%%%%%%%%%%%%
\begin{abstract}
We present a novel autonomous driving framework, DualAD, designed to imitate human reasoning during driving. DualAD comprises two layers: a rule-based motion planner at the bottom layer that handles routine driving tasks requiring minimal reasoning, and an upper layer featuring a rule-based text encoder that converts driving scenarios from absolute states into text description. This text is then processed by a large language model (LLM) to make driving decisions. The upper layer intervenes in the bottom layer's decisions when potential danger is detected, mimicking human reasoning in critical situations. Closed-loop experiments demonstrate that DualAD, using a zero-shot pre-trained model, significantly outperforms rule-based motion planners that lack reasoning abilities. Our experiments also highlight the effectiveness of the text encoder, which considerably enhances the model's scenario understanding. Additionally, the integrated DualAD model improves with stronger LLMs, indicating the framework's potential for further enhancement.
Code and benchmarks are available at \href{https://github.com/TUM-AVS/DualAD}{\texttt{github.com/TUM-AVS/DualAD}}.
\end{abstract}
%%%%%%%%%%%%%%%%%%%%%%%%%%%%%%%%%%%%%%%%%%%%%%%%%%%%%%%%%%%%%%%%%%%%%%%%%%%%%%%%
\vspace{-2mm}
\section{INTRODUCTION}
At the current stage of Autonomous Driving (AD), critical and rare scenarios, also called corner cases, are becoming one of the biggest challenges~\cite{chen2024end}. These cases often require a high level of reasoning ability. In response, researchers have been working to integrate reasoning capabilities into AD systems~\cite{shao2023reasonnet, wu2020motionnet, 9013045, teichmann2018multinet}. Recently, the emergence of large language models (LLMs) like GPT has provided researchers with another potential tool: leveraging the intelligence exhibited by these models to enhance the reasoning abilities required in autonomous driving. Some pioneering approaches that apply LLMs in autonomous driving have been developed~\cite{cui2023drivellm, xu2024drivegpt4, sima2023drivelm, dingrui2024esp}. However, while these approaches have shown LLMs' reasoning potential for autonomous driving, they haven't included close loop simulation, which is essential to validate a planner's performance. Also, recent approaches primarily focus on replacing current AD systems with LLMs instead of maximizing their full potential of the overall system. To fully maximize the potential of LLMs in autonomous driving systems, we need to explore strategies beyond merely replacing existing components. One promising approach is to draw inspiration from human driving styles.

Unlike power-intensive computers, the human brain has evolved to be computationally efficient, using roughly 15 watts~\cite{hofman2014evolution} to perform most daily tasks, including driving cars. Research shows that, for humans, complex driving scenarios are more cognitively demanding than simple ones~\cite{horberry2006driver}. Furthermore, human drivers can adapt their attention to meet the demands of different traffic conditions~\cite{kircher2017minimum, liu2021drivers}. 
These studies suggest that human drivers do not necessarily pay full attention to all driving situations; rather, they rely on simple causal relationships to handle regular scenarios. High-level reasoning abilities are used to supervise the environment in a non-intensive manner, intervening primarily when drivers encounter more cognitively demanding situations such as critical, risky, or challenging scenarios.

Building on these insights, we propose an AD framework that mimics human cognitive processes. Our core insight is that we can use LLMs as the reasoning module to check potential dangers and improve the current planners' performance. In this way, we can reduce the inference cost. Concretely, our contributions are twofold:

\begin{itemize} 
    \item We develop a rule-based text encoder to convert driving scenario into a format of text description. The experiments shown that with this text encoder, the LLMs tend to have a better understanding of the driving scenario and the integrated model can plan better.
    \item We introduce \textbf{DualAD}, a \textbf{Dual}-layer \textbf{A}utonomous \textbf{D}riving framework designed to replicate the human approach to driving by combining simple rule-based motion planning with an LLM for reasoning about desired velocity. Closed-loop experiments show that even with a weak, zero-shot LLM, our approach significantly improves performance of the rule-based planners.
\end{itemize}
\vspace{-0.25in}

\section{RELATED WORK}
\boldparagraph{Reasoning in Autonomous Driving}The task of integrating reasoning abilities into Autonomous Driving (AD) systems has been explored in various studies. Scene segmentation has been used to enhance models' semantic understanding~\cite{teichmann2018multinet, li2023mseg3d, sauerbeck2023camradepth} of the surrounding environment. However, these methods are perception-oriented and do not significantly enhance planning-oriented reasoning. Esterle et al.~\cite{esterle2018spatiotemporal} treat the task as a combinatorial problem, combining trajectory planning and maneuver reasoning, but complex scenarios pose a significant challenge for this method.
Multiple approaches have applied spatial and temporal modules to encode multimodal data to extend AD systems' reasoning abilities~\cite{shao2023reasonnet, hu2023planning, li2022bevformer, wu2020motionnet}. However, since these methods solely rely on data-driven approaches, they suffer from data-related issues such as covariate shift and domain adaptation~\cite{chen2024end}.
Kothawade et al.~\cite{kothawade2021auto} developed a rule-based model to encode the driving scenario into a sequence of text and use a predefined answer set to map the scenario description into a driving decision. However, the reasoning ability of this model is highly limited by the variety of the predefined answer set. In contrast, our approach leverages the flexibility of large language models to reason about complex driving scenarios.

\boldparagraph{Large Language Models in Autonomous Driving}A significant portion of the literature has focused on the application of Vision-Language Models (VLMs) and Large Language Models (LLMs) for driving tasks. Several approaches~\cite{nie2023reason2drive, xu2024drivegpt4, cui2023drivellm} utilize a visual encoder to parse driving images and build scene descriptions. While Chen et al.~\cite{chen2024driving} trained a scenario encoder to encode the agents' states. Some approaches~\cite{sima2023drivelm, ma2023dolphins} apply Vision-Language Models for planning to enhance scene understanding. \cite{ding2024holistic} created a dataset of instruction-response pairs to improve scene understanding in multimodal LLMs. GPT-Driver~\cite{mao2023gpt} utilizes text descriptions of scenarios as input, allowing the model to directly output planning trajectories. However, all the approaches mentioned above haven't conducted closed-loop simulation to validate the method. And recent work has emphasized the need to reconsider the metrics and evaluation methods for open-loop simulations~\cite{codevilla2018offline, zhai2023rethinking, li2024ego}, and closed-loop evaluation is regarded more correlated with driving quality.
Some works incorporate closed-loop simulations~\cite{wang2024drive, wen2023dilu}, albeit in limited and simple scenarios, which did not fully utilize the reasoning capabilities of LLMs.
Tian et al.~\cite{tian2024drivevlm} used VLMs to extract information such as weather and traffic signs to suggest driving speed limits, but dynamic agents were not considered during decision-making.  Shao et al.~\cite{shao2023lmdrive} trained a scenario encoder and a action decoder alongside a frozen LLM. However, their results are unstable and vary significantly across different scenarios.
Our proposed DualAD framework not only uses closed-loop evaluation to examine the method, but also tests the model in complex and critical scenarios.
 
\boldparagraph{Human-Like Reasoning Style in Driving} The idea of mimicking human cognitive processes in artificial intelligence has been a long-standing goal in the field~\cite{mnih2015human, silver2016mastering}. Recent approaches have aimed to create more holistic models that replicate broader aspects of human reasoning and adaptability~\cite{lake2017building}. For example, Kircher et al.\cite{kircher2017minimum} and Liu et al.\cite{liu2021drivers} explored how human drivers adjust their cognitive resources depending on the complexity of the driving task. Inspired by these insights, our work introduces a dual-layer framework that mirrors the human tendency to apply higher cognitive effort during complex or dangerous driving scenarios, thus enhancing the system's overall efficiency and safety.

\begin{figure}
    \centering
    \includegraphics[width=1\linewidth]{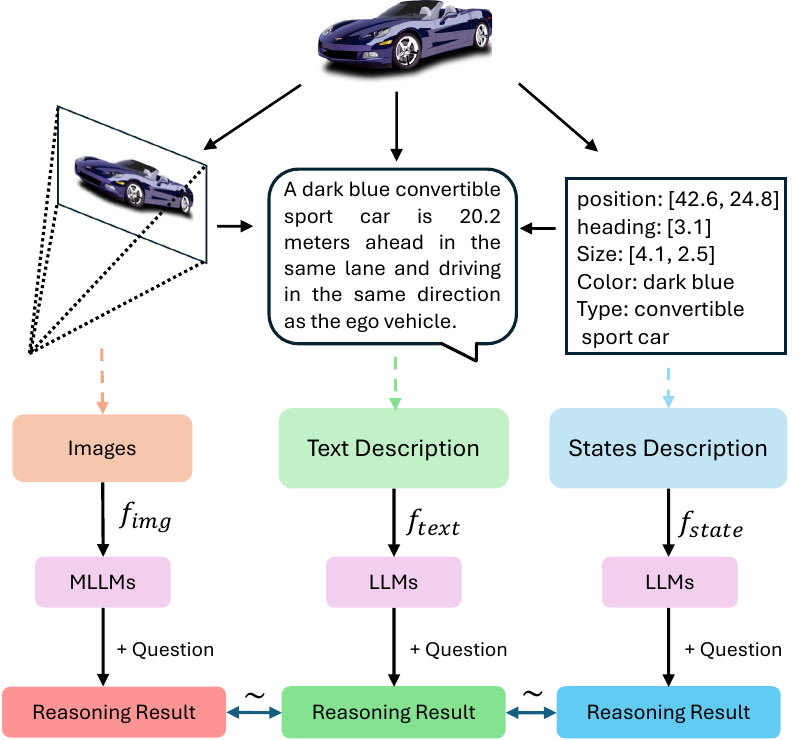}
    \vspace{-0.5em}
    \caption{Illustration of the reasoning processes comparing image and text modalities. The reasoning result of an image and the reasoning result of different types of the image related text description should be equivalent if the Platonic Representation Hypothesis~\cite{huh2024prh} holds.}
    \label{fig:text_and_states}
\vspace{-1.5em}
\end{figure} 

\section{METHODOLOGY}
The ultimate goal of our model is to produce a safe trajectory over $n$ seconds in each planning cycle. During each planning cycle, the planner receives a variety of inputs, which include tracking data of nearby objects, the current and historical kinematic states of the ego vehicle, traffic light information, high-definition (HD) maps, speed limits, and the specified route. We first present a rule-based encoder that converts the environment into a text description. Then we introduce the reference path planner and rule-based motion planners, followed by an explanation of how a large language model influences driving decisions.

\begin{figure*}[ht]
\centering
\includegraphics[width=2.07\columnwidth]{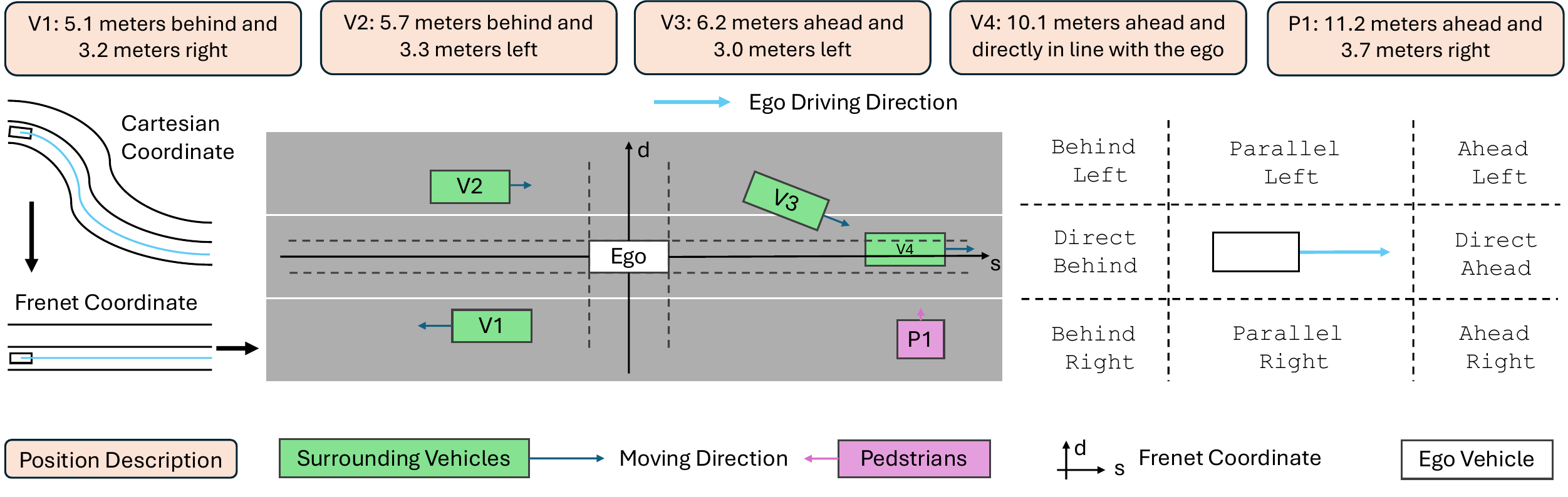}
\caption{The process of converting a driving scenario into a text description. The agents in the scenario are first transformed from Cartesian coordinates (in local frame) to Frenet coordinates using the reference path of the Motion Planner. Then, all agents are described in this local view based on their states through a rule-based system.}
\label{fig:text_descriptor}
\vspace{-0.25in}
\end{figure*}

\subsection{Upper Layer: Convert Driving Scenario into Text}
As illustrated in Fig.~\ref{fig:text_and_states}, \cite{huh2024prh} proposed that neural networks, when trained with different objectives on diverse data and modalities, converge towards a shared statistical model of reality. This suggests that reasoning results from different modalities (e.g., images, text, etc.) should be equivalent if the model is intensively trained on the corresponding forms. However, considering that the current LLM's capabilities are primarily trained on textual descriptions~\cite{achiam2023gpt}, the reasoning results derived from text are generally regarded as superior. Hence, instead of relying solely on tuples of exact numbers, such as position and velocity states, we aim to design this encoder to be able to give a text description that enable the large language model to achieve a deeper understanding of the semantics of driving scenarios.
As shown in Fig.~\ref{fig:text_descriptor}, we convert driving scenarios into text description. Each scenario $S$ comprises different types of agent, such as vehicles, pedestrians, traffic objects etc. $S$ is converted into a text description $D$ through the text encoder $f_\mathrm{text}$ as shown below,  
\begin{equation}
    D = f_\mathrm{text}(S)
\end{equation}
All agents are abstracted into states, which are tuples of agent's ID, position, orientation, speed and size. 
Using the reference path $P_{ref}$ planned by the path planner, we convert the pose $Pos_{cart} = (x, y, \theta_{\text{cart}})$ of each agent from Cartesian coordinates (in the local frame) to Frenet coordinates $Pos_{fren} = (s, d, \theta_{\text{fren}})$.
\begin{equation}
Pos_{fren} = Frenet(P_{ref}, Pos_{cart})
\end{equation}
Given an agent's pose in Frenet coordinates $(s, d, \theta_{\text{fren}})$, the following methodology is used to describe the agent's relative position and orientation:
\subsubsection{Longitudinal Position}
The longitudinal position $D_{lon}$ relative to the ego vehicle is determined as follows:
\begin{equation*}
\mathbox[red]{D_{lon}} =
\begin{cases} 
s \text{ meters ahead} & \text{if } s > 1, \\
|s| \text{ meters behind} & \text{if } s < -1, \\
\text{parallel with the ego} & \text{if } -1 \leq s \leq 1.
\end{cases}
\end{equation*}
\subsubsection{Lateral Position}
The lateral position $D_{lat}$ relative to the ego vehicle is determined as follows:
\begin{equation*}
\mathbox[green]{D_{lat}} =
\begin{cases} 
d \text{ meters left} & \text{if } d > 1, \\
|d| \text{ meters right} & \text{if } d < -1, \\
\text{directly in line with the ego} & \text{if } -1 \leq d \leq 1.
\end{cases}
\end{equation*}
\subsubsection{Orientation Description}
First, the agent's orientation $\theta_{\text{fren}}$ is normalized to fall within the range \([- \pi, \pi]\) with:
\begin{equation}
O_{norm} = (\theta_{\text{fren}} + \pi) \mod (2\pi) - \pi.
\end{equation}
The orientation of an agent is described as:
\begin{multline*}
\mathbox[violet]{D_{or}} = 
\begin{cases}
L_{or}[0]& \text{if} -\alpha \leq O_{norm} \leq \alpha, \\
L_{or}[1]& \text{if } O_{norm} \leq -\beta \text{ or } O_{norm} \geq \beta, \\
L_{or}[2]& \text{if } y \geq \gamma \text{ and } -\beta \leq O_{norm} \leq -\alpha, \\
L_{or}[2]& \text{if } y \leq -\gamma \text{ and } \beta \geq O_{norm} \geq \alpha, \\
L_{or}[3]& \text{otherwise.}
\end{cases}
\end{multline*}
with $\alpha$ is set to 0.06 rad, $\beta$ is set to 3.08 rad, $\gamma$ is equal to 1 and $L_{or}$ is the orientation description list as shown below,
\[
 \left[
\begin{array}{l}
\text{"$W$ in the same direction as the ego vehicle"}, \\
\text{"$W$ in the opposite direction of the ego vehicle"}, \\
\text{"$W$ towards the ego vehicle's planned trajectory"}, \\
\text{"$W$ away from the ego vehicle's planned trajectory"}
\end{array}
\right]
\]
For agents with speed \( v \geq 0.01 \) m/s (moving), the \text{$[W]$} variable in $L_{or}$ is set as "moving". For agents with speed \( v < 0.01 \) m/s (stationary), \text{$[W]$} is set to "facing".

\subsubsection{Final Description}
The final description combines the relative position and orientation description is represented as:
\begin{equation}
    {
    \color{gray}
    {\color{black}D}}{}= 
    {
    \color{gray}
    \overbracket[1pt]{\mathbox[blue]{\color{black}f_m}}^{\mathclap{\textsf{Description Format}}}}
    (
    {
    \color{gray}
    \underbracket[1pt]{\mathbox[red]{\color{black}D_{lon}}}_{\mathclap{\textsf{Longitudinal}}}}
    +
    {
    \color{gray}
    \overbracket[1pt]{\mathbox[green]{\color{black}D_{lat}}}^{\mathclap{\textsf{Lateral}}}}
    +
    {
    \color{gray}
    \underbracket[1pt]{\mathbox[violet]{\color{black}D_{or}}}_{\mathclap{\textsf{Orientation}}}}
    )
\end{equation}

\begin{figure}[h]
    \centering
    \includegraphics[width=1\linewidth]{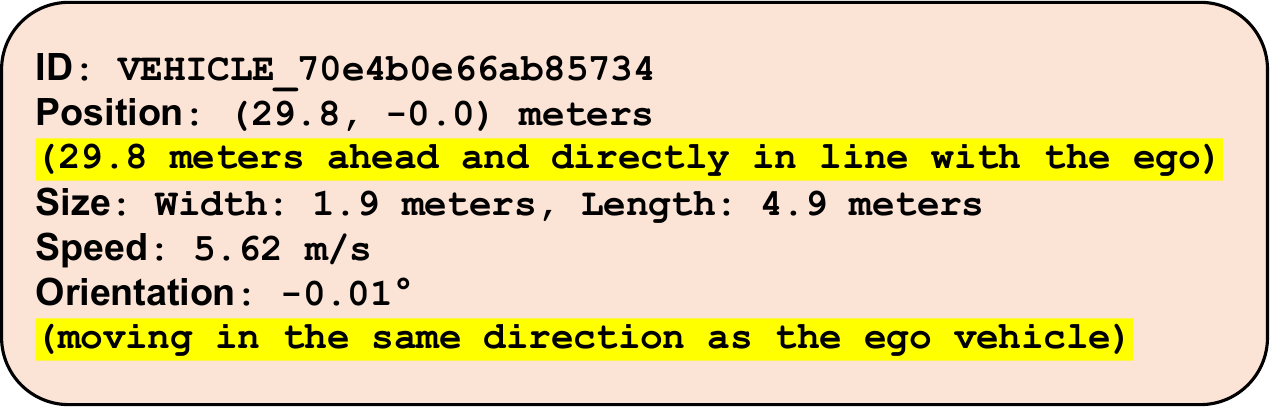}
    \vspace{-1.2em}
    \caption{An example of the text description of states for a vehicle.}
    \label{fig:text_description_eg}
\vspace{-2.5em}
\end{figure}

The format of the final description $f_m$ is shown below, 
\begin{equation*}
\mathbox[blue]{f_m} = 
\begin{cases} 
    \text{ID: agent\_id} \\
    \text{Position: } (x, y) \text{ meters } (D_{vp} + D_{hp}) \\
    \text{Size: }\text{Width: } w \text{ m, } \text{Length: } l \text{ m}\\
    \text{Speed: } v \text{ m/s} \\
    \text{Orientation: } \theta \text{ rad }  (D_{or})
\end{cases}
\end{equation*}
and a concrete example of the description is shown in Fig.~\ref{fig:text_description_eg}.

\vspace{-1.5mm}

\subsection{Bottom Layer: Rule-based planners}
In this section we briefly introduce the different rule-based planners used in the experiments with our DaulAD framework.
\boldparagraph{Intelligent Driver Model}The Intelligent Driver Model (IDM)~\cite{treiber2000congested} is a basic planning method that is used to provide a reference for the ego-vehicle's planning. IDM calculates the path based on the road centerline and adapts the velocity along that path during simulation. Based on the speed $v$, and the longitudinal distance $s$ to the vehicle ahead on the centerline, IDM repeatedly applies the following rule to determine the acceleration along the path:
\begin{equation}
    \frac{\diff v}{\diff t} = a \Bigl(1 - \Bigl(\frac{v}{v_0}\Bigr)^{\delta} - \Bigl(\frac{s^*}{s}\Bigr)^2 \Bigr)
\end{equation}
where $a$ is the acceleration limit, $v_0$ is the target speed, $s^*$ is the safety distance, and $\delta$ is an exponent. These values are chosen manually. The behavior of the model is as follows. It accelerates the ego vehicle to the target speed $v_0$ or decelerates if it's too close to the vehicle ahead (at a distance $s^*$).

\boldparagraph{Lattice Planner}A lattice planner used in~\cite{Huang_2023_ICCV} discretizes the continuous search space into a regular grid or lattice, where each point on the grid represents a potential state or position that the agent can occupy. The primary objective of a lattice planner is to find a feasible and optimal path from a start state to a goal state while satisfying the system's motion constraints and avoiding obstacles.

The essence of a lattice planner can be mathematically described by considering the state space \( S \), which is discretized into a lattice grid \( \mathcal{L} \). The planner's goal is to find a sequence of states \( \{s_0, s_1, \dots, s_n\} \subset \mathcal{L} \) where \(s_0 = s_{\text{start}}, s_n = s_{\text{goal}}\) that minimizes a cost function \( J \), subject to the dynamic constraints of the system.
The optimal path can be represented by solving the following optimization problem:
\begin{equation}
\min_{s_0, s_1, \dots, s_n} \sum_{i=0}^{n-1} c(s_i, s_{i+1}) \quad s.t. \quad \Phi(s_i, s_{i+1}) \leq 0
\end{equation}
where \( s_i \in \mathcal{L}, \text{and } \Phi(s_i, s_{i+1}) \leq 0\), \( c(s_i, s_{i+1}) \) is the cost of moving from state \( s_i \) to state \( s_{i+1} \), \( \Phi \) represents the motion constraints of the system and \( s_{\text{0}} \) and \( s_{\text{n}} \) are the initial and goal states, respectively.

\begin{figure*}[ht]
\centering
\includegraphics[width=2.05\columnwidth]{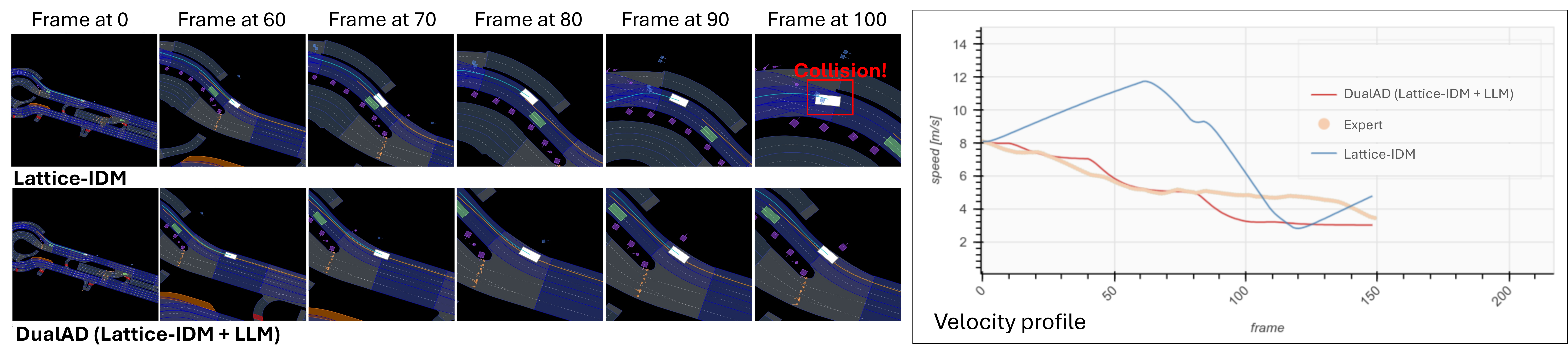}
\caption{Performance comparison between DualAD and Lattice-IDM on an example simulated with reactive environment.}
\label{fig:performance_compare}
\vspace{-0.0in}
\end{figure*}

\begin{table*}[t]
\vspace{6pt}
\begin{center}
\setlength{\tabcolsep}{15pt}
\renewcommand{\arraystretch}{1.2}
\small
\begin{tabular}
{cccccccc}
\toprule
\multicolumn{2}{c}{Planners} & \multicolumn{2}{c}{Hard-55} & \multicolumn{2}{c}{Super-Hard-24}   \\ \midrule
Type & \multicolumn{1}{l|}{Method} & NR-CLS $\uparrow$ & \multicolumn{1}{c|}{R-CLS $\uparrow$} & NR-CLS $\uparrow$ & \multicolumn{1}{c}{R-CLS $\uparrow$}  \\ \midrule
\fadedtext{Expert} & \multicolumn{1}{l|}{\fadedtext{Log-replay}} & \fadedtext{76.13} & \multicolumn{1}{c|}{\fadedtext{58.32}} & \fadedtext{86.52} & \multicolumn{1}{c}{\fadedtext{48.70}}  \\ \midrule
\multirow{4}{*}{Rule-based} 
& \multicolumn{1}{l|}{\mathbox[red]{\text{IDM}}~\cite{treiber2000congested}} & 50.12 & \multicolumn{1}{c|}{34.56} & 53.90 & \multicolumn{1}{c}{20.73}   \\
& \multicolumn{1}{l|}{\mathbox[violet]{\text{Lattice-IDM}}~\cite{treiber2000congested, Huang_2023_ICCV}} & \textcolor{orange}{52.36} & \multicolumn{1}{c|}{\textcolor{orange}{39.76}} & \textcolor{orange}{57.01} & \multicolumn{1}{c}{\textcolor{orange}{33.83}}  \\
& \multicolumn{1}{l|}{\mathbox[green]{\text{Frenetix}}~\cite{trauth2024frenetix} (Ours)} & \textbf{78.77} & \multicolumn{1}{c|}{77.80} & \textbf{76.77 }& \multicolumn{1}{c}{73.08}  \\
 & \multicolumn{1}{l|}{\text{PDM-Closed}~\cite{dauner2023parting}} & 58.33 & \multicolumn{1}{c|}{35.15} & 51.26 & 7.57 \\ \midrule
 \multirow{2}{*}{Learning-based} 
  & \multicolumn{1}{l|}{UrbanDriver~\cite{scheel2022urban}} & 42.21 & \multicolumn{1}{c|}{30.92} & 52.50 & \multicolumn{1}{c}{40.93}  \\ 
 & \multicolumn{1}{l|}{PlanTF~\cite{cheng2024rethinking}} & 62.44 & \multicolumn{1}{c|}{53,60} & 69.83 & \multicolumn{1}{c}{51.30}  \\ 
 \midrule
\multirow{3}{*}{Hybrid}  
& \multicolumn{1}{l|}{DualAD (\mathbox[red]{\text{IDM}}+ LLM)} & 49.37 & \multicolumn{1}{c|}{40.25} & 53.58  & \multicolumn{1}{c}{24.67}   \\ 
& \multicolumn{1}{l|}{DualAD (\mathbox[violet]{\text{Lattice-IDM}}+ LLM)} & \textbf{\textcolor{purple}{60.25}} & \multicolumn{1}{c|}{\textbf{\textcolor{purple}{57.31}}} & \textbf{\textcolor{purple}{57.61}} & \multicolumn{1}{c}{\textbf{\textcolor{purple}{46.03}}}  \\ 
& \multicolumn{1}{l|}{DualAD (\mathbox[green]{\text{Frenetix}}+ LLM)} & 75.55 & \multicolumn{1}{c|}{\textbf{77.86}$^\dagger$} & 76.09 & \multicolumn{1}{c}{\textbf{73.18}$^\dagger$}   \\ 

\bottomrule
\end{tabular}
\end{center}
\vspace{-0.6em}
\caption{Comparison with state-of-the-arts. The LLM mainly used in these experiments is GLM-4-Flash.\\
\textit{\footnotesize $^\dagger$ indicates the LLM used in the related experiment is GPT-4o.}}
\label{tab:sota}
\vspace{-4mm}
\end{table*}

\boldparagraph{Frenetix Motion Planner}Frenetix Motion Planner~\cite{trauth2024frenetix} uses a sampling approach to improve comfort, safety, and accuracy for trajectory planning in complex environments. 
In each iteration, a number of kinematically feasible polynomial trajectories $\mathcal{T}$ is spanned from the current ego vehicle's position. The trajetories vary in their final lateral displacement, velocity as well as their planning horizon. 
For each trajectory $\xi \in \mathcal{T}$, the costs are calculated based on a cost function which is related to different factors $J_\mathrm{i}(\xi)$ with weighting $\omega_\mathrm{i}$. 
\begin{equation}
    J_\mathrm{sum}(\xi|f_{\xi}) = \sum_{i=1}^{n} \omega_\mathrm{i} \cdot J_\mathrm{i}(\xi)
\end{equation}
The cost terms encompass accelerations for a comfortable trajectory selection, the deviation from the desired velocity as well as the distance to the desired global route and the risk of colliding with observed agents. This combination of factors ensures an efficient and comfortable progress while ensuring safety. 
The trajectory with the lowest cost  is then chosen as the optimal trajectory ${\xi}_{ optimal}$.
\vspace{-1.5mm}
\subsection{Planning Intervention by the LLM}
After the text-based scenario description is fed into the LLM, we update the desired velocity based on the LLM output. The reasoning result $I_{R}$ is shown below, 
\begin{equation}
    I_{R} = LLM(f_\mathrm{text}(S)) 
\end{equation}
The reasoning result includes the suggested driving speed limit. The allowed speed suggestions range from 0 to 15 m/s.
If the rule-based planner plan a speed $v_{rule}$ that is higher than the LLM's suggestion, then the decision of LLMs will be used to overwrite the driving decision of the rule-based planner as shown below,
\begin{equation}
v_{d} = 
\begin{cases}
I_{R}[\text{"speed"}] & \text{if } I_{R}[\text{"speed"}] \leq v_{rule}\\
v_{rule}& \text{otherwise.}
\end{cases}
\end{equation}

\section{Experiments} 
\boldparagraph{Dataset and Simulation}NuPlan~\cite{caesar2021nuplan} serves as the closed-loop ML-based benchmark for autonomous vehicle planning. The dataset includes 1300 hours of driving data recorded across four different cities. We conducted all evaluations on the public nuPlan mini set, which contains over 2,000 diverse scenarios. We also explored two different strategies for selecting scenarios:
\textbf{Hard-55:} In this strategy, scenarios are filtered out by the score of the R\_CLS metric for the IDM. The 55 scenarios with the worst scores are chosen. Since the IDM achieves an intermediate performance in closed-loop scenarios, the filtered result scenarios are regarded relatively hard.
\textbf{Super-Hard-24:} This strategy is designed to test the planner's ability to handle even more challenging scenarios. We ran 2000 scenarios in total using the PDM-Closed~\cite{dauner2023parting} planner, which is current state-of-the-art as it ranks first in the planning leaderboard. The 24 scenarios with the worst scores are chosen and are regarded more challenging.

Our simulation environment is nuPlan's closed-loop simulator. 
Each simulation consists of a 15-second rollout at a frequency of 10 Hz. The simulator uses an LQR controller for tracking the planned trajectory. Background traffic behavior is influenced by the simulation mode, which can be either non-reactive (log-replay) or reactive.
\vspace{1mm}

\boldparagraph{Metrics}The evaluation metrics used in this study are the official ones provided by nuPlan~\cite{caesar2021nuplan}, including the open-loop score (OLS), non-reactive closed-loop score (\textbf{NR-CLS}), and reactive closed-loop score (\textbf{R-CLS}). While Codevilla et al.~\cite{codevilla2018offline} argued that open-loop evaluation is not necessarily correlated with driving quality, we use only closed-loop evaluation. The NR-CLS and R-CLS are calculated using the same methodology, with the key difference being that R-CLS includes background traffic control using the IDM~\cite{treiber2000congested} during the simulations. The closed-loop score is a composite score derived from a weighted combination of several factors, including similarity to human driving, vehicle dynamics, and goal achievement etc. The score ranges from 0 to 100.

\boldparagraph{Zero-shot Large Language Models}We selected two large language models to build the reasoning module, which uses LLMs for inference. The first is the freely available GLM-4-Flash, a member of the General Language Model (GLM) family~\cite{glm2024chatglm}. The second model is GPT-4o~\cite{OpenAI2024}, which is considered one of the most advanced LLM to date. It is important to note that, unless otherwise stated, we use GLM-4-Flash as the default reasoning module. GPT-4o is used only in experiments involving comparisons of different levels of LLM influence and in ablation studies.

\vspace{0.8mm}
We compare the performance of different rule based planners with and without the Reasoning of LLMs and compare them against the current state-of-the-art.
\vspace{-1.8mm}

\subsection{Rule-based Model with and without LLM Reasoning}
\boldparagraph{IDM}We evaluated the performance of IDM with and without the integration of an LLM. Our findings indicate that incorporating the LLM significantly enhances IDM's effectiveness. As shown in Table.~\ref{tab:sota}, without the reasoning module, IDM's performance is relatively low across both benchmarks for reactive (NR-CLS) and nonreactive (R-CLS) simulations. However, when the LLM is integrated, the R-CLS score in the Hard-55 benchmark increases by 16\%, and in the SuperHard-24 benchmark, the R-CLS score saw an improvement of nearly 20\%. Notably, the R-CLS score in the Hard-55 benchmark surpasses most of rule-based and learning-based methods, including PDM-Closed~\cite{dauner2023parting}, which is regarded as the state-of-the-art. In contrast, the NR-CLS score did not show a significant increase in either benchmark.

\boldparagraph{Lattice-IDM}Next, we evaluated the Lattice-IDM approach, both with and without LLM support. The Lattice-IDM planner combines the traditional IDM model with a lattice-based path planning approach.
As indicated in Table.~\ref{tab:sota}, without LLM support, although the Lattice-IDM achieves higher R-CLS scores on the Hard-55 benchmark than IDM, PDM-Closed and UrbanDriver~\cite{scheel2022urban}, the number is still relatively low. 
Additionally, the Lattice-IDM's overall performance across both benchmarks is also very low.
While with the assistance of the LLM, however, the scores in R-CLS not only surpass most of rule-based and all the learning-based methods on the Hard-55 benchmark, but also exceed those of DualAD (IDM + LLM). And the scores on other metrics are also improved significantly. 
Notably, when compared to not using LLMs, the scores of DualAD with the Lattice-IDM planner increased on the Hard-55 R-CLS benchmark by \textbf{44\%}, and on the SuperHard-24 benchmark by \textbf{36\%}. As shown in Fig.~\ref{fig:performance_compare}, this example demonstrates that DualAD with the help of LLM, not only avoids collisions by reducing speed in advance but also presents a smoother velocity profile.

\boldparagraph{Frenetix}Then we tested the Frenetix planner with and without the assistance of the LLM. Without LLM, Frenetix outperforms all rule-based and learning-based methods. And with the assitance of LLM, scores in R-CLS on both benchmarks is still enhanced, the improvements are not very evident. And DualAD with Frenetix scores similarly in NR-CLS on both benchmarks compare to without using LLM.

\begin{table}[t]
\begin{center}
\setlength{\tabcolsep}{9pt}
\renewcommand{\arraystretch}{1.2}
\small
\begin{tabular}{cc|ccccc}
\toprule
\multirow{2}{*}{LLM} & \multirow{2}{*}{Tasks}  &  \multicolumn{2}{c}{Planners} \\ 
&  & IDM & Lattice-IDM \\ \midrule
\multirow{2}{*}{GPT-4o} & \mathbox[green]{\text{NR-CLS}}$\uparrow$ & 66.09 & \textbf{76.72}  \\
                        & \mathbox[red]{\text{R-CLS}}$\uparrow$ & \textbf{31.91} & \textbf{44.88}  \\ \midrule
\multirow{2}{*}{GLM-4-flash} & \mathbox[green]{\text{NR-CLS}}$\uparrow$ & {69.31}$^\ast$ & 64.26$^\dagger$  \\
                        & \mathbox[red]{\text{R-CLS}}$\uparrow$ & 19.04$^\S$ & 22.07$^\dagger$  \\  \midrule
\multirow{2}{*}{No LLM} & \mathbox[green]{\text{NR-CLS}}$\uparrow$ & 68.88 & 69.69  \\
                        & \mathbox[red]{\text{R-CLS}}$\uparrow$ & 6.81 & 13.52 \\ 
                        \bottomrule
\end{tabular}
\end{center}
\vspace{-2mm}
\caption{Results of the influence of different reasoning levels of LLMs on the integrated DualAD.
\textit{\footnotesize $^\ast$, $^\dagger$ and $^\S$ indicate proportions of failed simulations with related methods are 10\%, 20\% and 30\%, respectively.} }
\label{tab:llm_level}
\vspace{-4mm}
\end{table}

\vspace{-0.6mm}
\subsection{Different Levels of LLMs}\label{subsec:diff_level}
We randomly choose ten scenarios from the SuperHard-24 benchmark to explore how different reasoning levels of LLMs can influence the integrated DualAD's performance.
As shown in Table~\ref{tab:llm_level}, the performance of the integrated model is directly influenced by the level of the LLMs used. We used two different levels of LLMs: GPT-4o (considered the most advanced LLM currently available) and GLM-4-flash (a relatively less powerful model). The results demonstrate that using a stronger LLM like GPT-4o in Lattice-IDM led to a nearly 20\% improvement in NR-CLS and over a 100\% increase in R-CLS for the integrated DualAD model. Additionally, GPT-4o not only enhanced performance but also provided more stable outputs. In our experiments, 20\% of simulations using GLM-4-flash failed because the model's output did not conform to the required format. In contrast, all simulations using GPT-4o were successful.
\vspace{-0.6mm}
\subsection{Ablation Study}
\vspace{-0.6mm}
We used the same benchmark setting as Subsection~\ref{subsec:diff_level} to evaluate the impact of text encoder on the overall system performance, we conducted an ablation study using GPT-4o. Table \ref{tab:ablation_study} illustrates that omitting the text encoder adversely impacts the performance of the Large Language Model. In experiments where Lattice-IDM is integrated with the LLM, the exclusion of the text encoder leads to a reduction of over 11\% in R-CLS scores and more than 7\% in NR-CLS scores.

\section{DISCUSSION}
We compared DualAD's performance with several state-of-the-art planners. It is notable that \textbf{PDM-Closed}~\cite{dauner2023parting}, a rule-based method that integrates IDM with different hyperparameters, achieves the highest score on the official leaderboard. \textbf{PlanTF}~\cite{cheng2024rethinking} achieves the best performance among all learning-based planners, and \textbf{UrbanDriver}~\cite{scheel2022urban} is a vectorized Transformer-based planner. However, DualAD with the Lattice-IDM planner outperforms PDM-Closed and UrbanDriver on all benchmarks. Additionally, DualAD with the Lattice-IDM planner achieves similar performance and even outperforms PlanTF in R-CLS on the Hard-55 benchmark.
Furthermore, with a large language model's help, DualAD with different rule-based models, including IDM and Lattice-IDM, significantly outperforms their original models. Most importantly, we achieved these improvements using only a relatively weak LLM, \textbf{GLM-4-Flash}, by simply adjusting the desired speed of the vehicle.
All of these improvements demonstrate that the DualAD framework, which combines rule-based methods with LLMs, can contribute to more reliable autonomous driving, particularly in critical scenarios.

We should also notice that \textbf{Frenetix}~\cite{trauth2024frenetix} is a well-designed planner, which is a complex system including modules for sampling, kinematic checks, prediction, and planning. Although the improvements for Frenetix are not as evident, we can observe significant improvements in weaker models. Since the reasoning ability of weak planners is minimal compared to LLMs, integrating LLMs leads to notable enhancements. However, because Frenetix already incorporates reasoning capabilities through its complex system of modules, its reasoning ability is relatively comparable to that of an LLM, and thus the LLM does not further improve its performance. One possible reason is that the current benchmark is not critical enough to challenge Frenetix's reasoning ability.

\begin{table}[t]
\begin{center}
\setlength{\tabcolsep}{6pt}
\renewcommand{\arraystretch}{1.2}
\small
\begin{tabular}{ccccccc}
\toprule
 & Text-Encoder & LLM & NR-CLS $\uparrow$ & R-CLS $\uparrow$ \\ \midrule
\multirow{3}{*}{IDM} 
 & \cmark & \cmark & 66.09 & \textbf{31.91} \\
 & \xmark & \cmark & 67.08 & 18.64 \\
 & \xmark & \xmark & 68.88 & 6.81 \\ \midrule
\multirow{3}{*}{Lattice-IDM} 
 & \cmark & \cmark & \textbf{76.72} & \textbf{44.88} \\
 & \xmark & \cmark & 71.01 & 39.78 \\
 & \xmark & \xmark & 69.69 & 13.52 \\ 
 \bottomrule
\end{tabular}
\end{center}
\vspace{-0.6em}
\caption{Results of ablation study regarding the text-encoder and LLM. The LLM used in this experiment is GPT-4o.}
\label{tab:ablation_study}
\vspace{-4mm}
\end{table}

\section{CONCLUSION}
This paper introduced DualAD, an autonomous driving framework that integrates a rule-based motion planner with a large language model to enhance decision-making in complex driving scenarios. DualAD effectively mimics human reasoning by employing the LLM for high-level reasoning in critical situations, while relying on the rule-based planner for regular tasks. The closed-loop experiments validated the framework's stable capability, showing that DualAD improves performance in challenging scenarios compared to other planners. The incorporation of a text encoder, which converts driving scenarios into a text format, is important for the framework's success.
\\
\boldparagraph{Limitation and Future Work}DualAD's text descriptions currently focus solely on surrounding agents, omitting crucial map details like lanes and sidewalks. Integrating this map information could enhance the LLM's understanding of the driving environment. Additionally, the reasoning module's driving decisions lack steering direction, which could be explored to maximize the LLM's reasoning abilities. Furthermore, the LLM only processes one frame at a time, which results in lacking historical data. Another future work can be feeding it multiple frames could provide temporal context.
%%%%%%%%%%%%%%%%%%%%%%%%%%%%%%%%%%%%%%%%%%%%%%%%%%%%%%%%%%%%%%%%%%%%%%%%%%%%%%%%
\newpage
%\section*{ACKNOWLEDGMENT}

\bibliographystyle{IEEEtran}
\bibliography{ref}

\end{document}